%% file: bmvc_final.tex
\documentclass{bmvc2k}
\usepackage{times}
\usepackage{epsfig}
\usepackage{graphicx}
\usepackage{amsmath}
\usepackage{amssymb}
\usepackage[linesnumbered, lined, ruled]{algorithm2e}
\usepackage{tabularx}
\usepackage{subfigure}
\usepackage{caption}
\usepackage{color}

\newcommand{\varA}[1]{{\operatorname{#1}}}
\title{Generative OpenMax\\ for Multi-Class Open Set Classification}

\addauthor{Zongyuan Ge$^{\star}$}{zongyuan@au1.ibm.com}{1}
\addauthor{Sergey Demyanov$^{\star}$}{sergey@demyanov.net}{1}
\addauthor{Zetao Chen}{chenze@ethz.ch}{2}
\addauthor{Rahil Garnavi}{rahilgar@au1.ibm.com}{1}

\addinstitution{
 IBM Research\\
 204 Lygon Street\\
 Melbourne, Australia
}
\addinstitution{
 Vision for Robotics Lab\\
 ETH Zurich\\
 Zurich, Switzerland
}

\runninghead{Student, Prof, Collaborator}{BMVC Author Guidelines}

\begin{document}
\maketitle

\input{contrib/abstract}
\input{contrib/intro}
\input{contrib/background}
\input{contrib/method}
\input{contrib/exp}
\input{contrib/conclusion}

\bibliography{egbib}
\end{document}

%% file: contrib/abstract.tex
\begin{abstract}
We present a conceptually new and flexible method for multi-class open set classification.
Unlike previous methods where unknown classes are inferred with respect to the feature or decision distance to the known classes, our approach is able to provide explicit modelling and decision score for unknown classes. The proposed method, called Generative OpenMax (G-OpenMax), extends OpenMax by employing generative adversarial networks (GANs) for novel category image synthesis. 
We validate the proposed method on two datasets of handwritten digits and characters, resulting in superior results over previous deep learning based method OpenMax
Moreover, G-OpenMax provides a way to visualize samples representing the unknown classes from open space.
Our simple and effective approach could serve as a new direction to tackle the challenging multi-class open set classification problem. 

%

\end{abstract}

%% file: contrib/intro.tex
\section{Introduction}
The computer vision community has progressively improved object classification performance in recent years. Driven by recent research in machine learning, the best performing system can achieve impressive results on both small and large scale multi-class classification tasks~\cite{he2015deep,he2017mask,szegedy2016inception,krizhevsky2012imagenet,russakovsky2015imagenet}.    
Most commonly, these recognition tasks are defined under a closed-set setting, i.e. all testing samples belong to one of the known classes in the training set. 
However, that is not the ideal setting to train a machine targeting real world applications. 
The most comprehensive image dataset these days, ImageNet~\cite{Deng2009} only contains 1000 classes. 
Considering only the number of animal species on the earth, the number of classes would exceed 1.5 million~\cite{may1988many} animal species on the earth, not to mention other general object categories. 

To tackle this challenge, open set classification topic has recently attracted much attention~\cite{scheirer2013toward,   scheirer2014probability,jain2014multi,bendale2015towards,bendale2016towards}.
Open set problem assumes that for those test objects who do not belong to any known classes (i.e. classes the model has been seen before and trained on), the classifier must correctly identify them as unknown class, as opposed to false classification to one of the known classes.
Multi-class open set classification is challenging because it requires the correct probability estimation of all known classes, together with simultaneous precise predicting of unknown classes. 
Given this,~\cite{scheirer2013toward} defines open set classification as a problem of balancing known space (specialization) and unknown open space (generalization) of the model (see Fig.~\ref{fig:openspace}). 
Formalization for open space risk is considered as the relative measure of open space compared to the overall measure space:
\begin{equation}
R_{open} = \frac{Unknown\; Space}{(Known\; Space + Unknown\; Open\; Space)}
\end{equation}

Several recent solutions to multi-class open set problem convert it into a score calibration task~\cite{niculescu2005predicting,smola2000advances,scheirer2011meta}. 
Those methods model the known class distribution using a parametric model from which the posterior probabilities of a test sample are computed.
The unknown class probabilities are then estimated through statistical modelling, specially Extreme Value Theory (EVT)~\cite{smith1990extreme}. 
However, those models do not assume any prior information about open space.
In essence, the ``open space risk'' is only estimated in decision space rather in pixel space.

Our proposed method, called Generative OpenMax (G-OpenMax), extends OpenMax~\cite{bendale2016towards} by providing explicit probability estimation over unknown categories. This is done by using generative adversarial networks (GANs)~\cite{goodfellow2014generative} to generate synthetic samples from unknown classes. The synthetic samples are generated from mixture distributions of known classes in latent space, which leads to plausible representation with respect to the known classes domain.
Explicit representation of unknown classes enables the classifier to locate the decision margin with the knowledge of both known and unknown samples, thus resulting in better balance of open space risk and known space for the multi-class open set model. The contributions of this paper can be summarized as:


\begin{figure*}
\begin{center}
\includegraphics[width=1\linewidth]{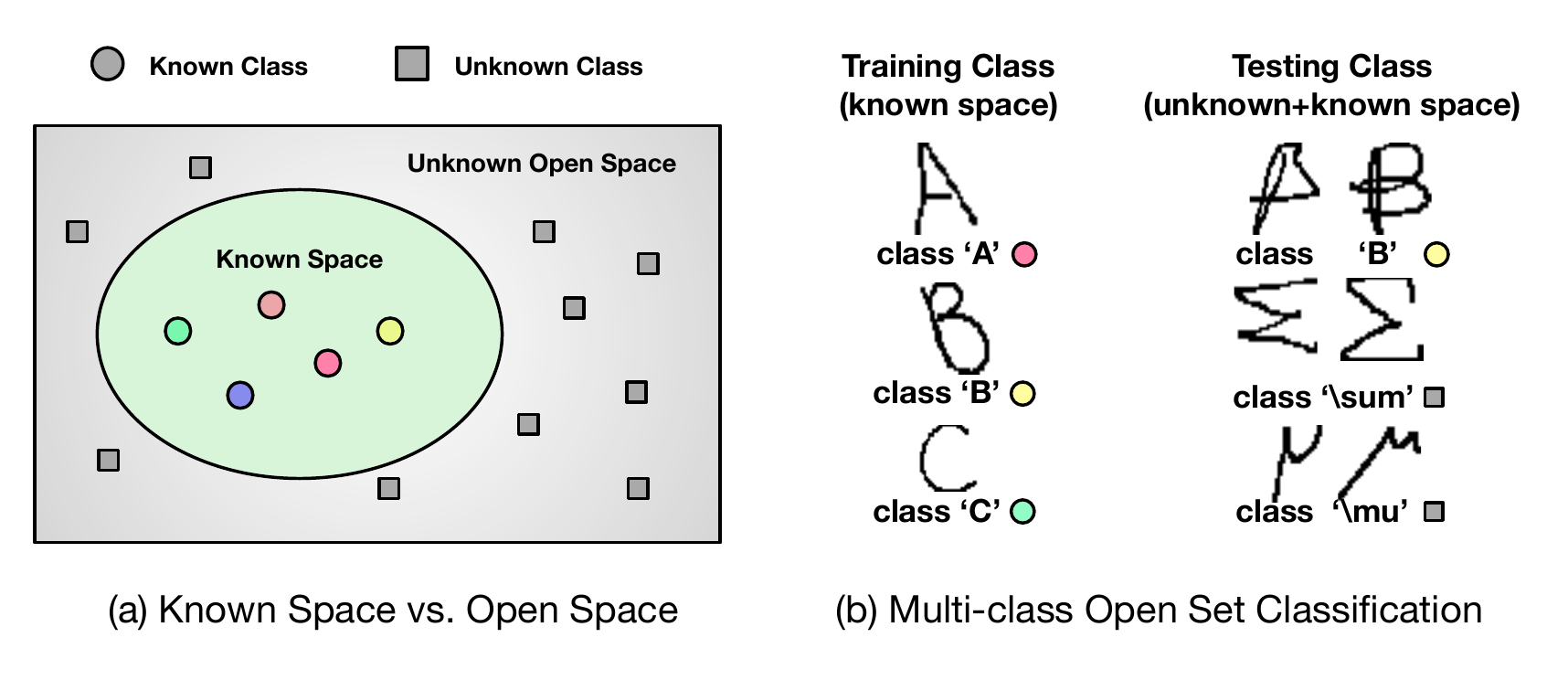}
\end{center}
   \caption{Figure (a) illustrates the relationship between the known space and unknown open space. Green area defines known space containing various known classes (coloured circles). Gray area represents open space and consists of unknown classes (grey rectangulars). Figure (b) illustrates the open set setting by showing classes from train/test phase, where classes such as ``$\mu$'' and ``$\sum$'' do not belong to the training set, hence unknown to the classifier during training (images are from HASYv2 dataset~\cite{thoma2017hasyv2}).}
\label{fig:openspace}
\end{figure*}

\begin{enumerate}
\item We propose a new method G-OpenMax to tackle the challenging multi-class open set classification. Our proposed method enables non-pseudo probability estimation and visualization over both known and unknown classes. 
\item We modify the training procedure of conditional GAN and mix several prior known classes distributions to generate plausible and domain adapted synthetic unknown samples.
\item Comprehensive experimental analysis have been conducted to evaluate the effectiveness and limitations of the proposed method on small scale (10 classes) and large scale (95 classes) openness problems. 
\end{enumerate}



%% file: contrib/background.tex
\section{Background}
\noindent\textbf{Openset set:} Early attempts to solve open set classification task involved adapted closed-set classifiers.
Scheirer et al.~\cite{scheirer2013toward} proposed ``1-vs-Set Machine'' which detects an unknown class by exploiting a decision space from the marginal distances of a binary SVM classifier. Later Scheirer et al.~\cite{scheirer2014probability} proposed compact abating probability (CAP) model which extended ``1-vs-Set Machine'' to non-linear $\varA{W-SVM}$ for multi-class open set scenario. 
Meanwhile, there were a few works exploring non-close-set methods for open set tasks.
Jain et al.~\cite{jain2014multi} introduced $\varA{P_{I}-SVM}$ to leverage Statistical Extreme Value Theory (EVT) and multi-class SVM scores to robustly estimate the posterior probability of known and unknown classes. 
Bendale et al.~\cite{bendale2015towards} offered the definition of ``open world recognition'' and proposed Nearest Non-Outlier (NNO) algorithm to actively detect and learn new classes to the model. 
Recently, driven by the effectiveness of deep networks, Bendale and Boult~\cite{bendale2016towards} adapted deep learning framework for open set recognition. Combining the concept of penultimate layer with meta recognition~\cite{scheirer2011meta}, they proposed OpenMax which enables basic deep network to reject either ``fooling'' or unknown open set classes.  

\noindent\textbf{Generative Adversarial Learning:} There are various approaches for generative models. The most prominent are variational autoencoder (VAE)~\cite{kingma2013auto} and generative adversarial networks (GAN)~\cite{goodfellow2014generative}, among which, GAN is a powerful model of learning arbitrarily complex data distributions. 
GAN is formulated on a game-theoretic using the philosophy of competition between two networks for training an image synthesis model. 
Recent work shows GAN is able to produce highly plausible images on both gray-scale and RGB images~\cite{radford2015unsupervised}.
Many improvements have been proposed to enhance the overall performance for training GANs~\cite{arjovsky2017wasserstein,salimans2016improved}.
GAN has a wide range of applications including, sketch/text to image synthesis~\cite{creswell2016adversarial}, image to image translation~\cite{isola2016image}, image editing~\cite{perarnau2016invertible}, super resolution~\cite{ledig2016photo} and video prediction~\cite{finn2016unsupervised}, etc. 
To the best of our knowledge, this is the \textbf{first} time generative model like GAN has been used for open set classification task.

%% file: contrib/method.tex
\section{Methodology}
Generative OpenMax (G-OpenMax) is conceptually simple. 
While OpenMax estimates the pseudo probability of unknown class by aggregating calibrated scores from known classes,  our proposed G-OpenMax, which is an intuitive solution, directly estimates the probability of unknown class. This is done by using synthetic images as an extra training label apart from known labels.   
Therefore, the main challenge is the following: how to generate plausible samples such that they meet the criteria of being distinct from the known classes, while they are well represent for the open space.
In Sec.~\ref{sec:g-openmax} and Sec.~\ref{sec:syn-samples}, we introduce the key elements of G-OpenMax, including how to synthesis unknown samples and post-processing to select suitable cases for open set classifier training.

\begin{figure*}
\begin{center}
\includegraphics[width=\linewidth]{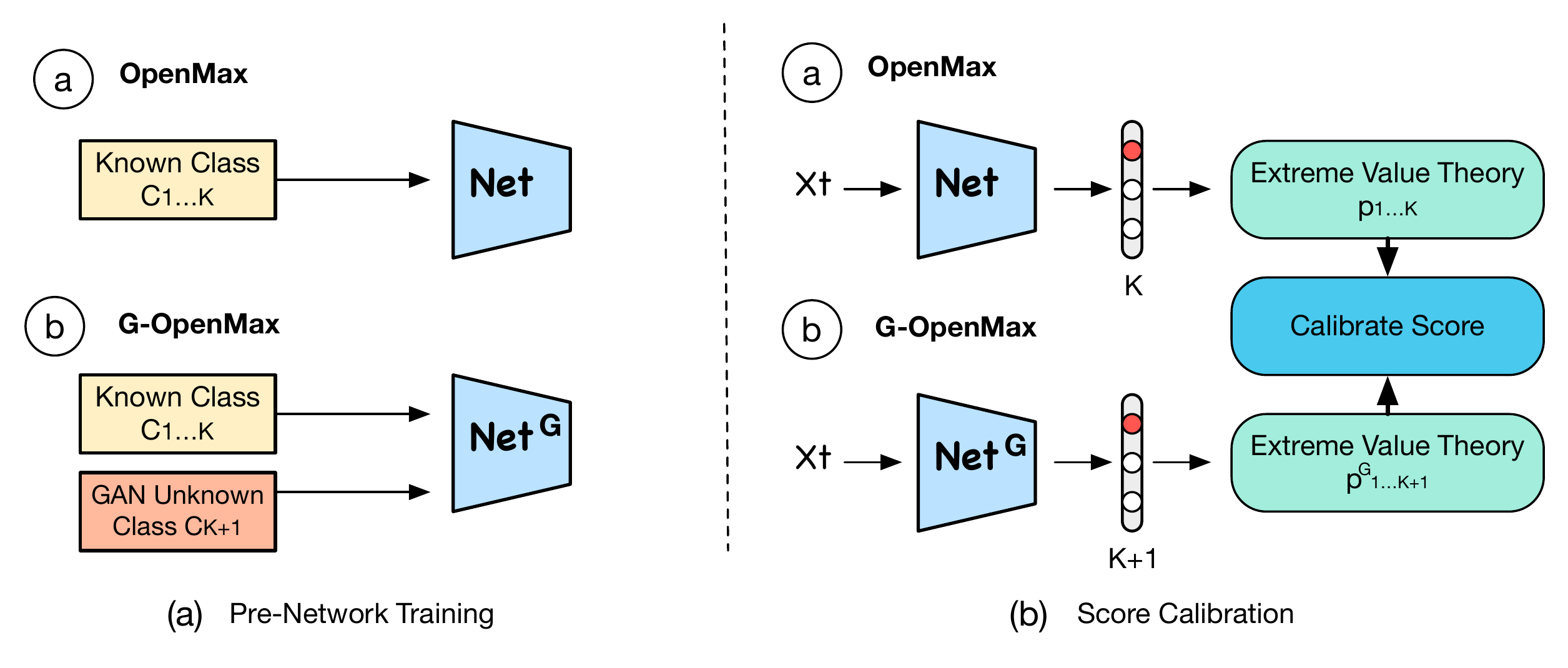}
\end{center}
   \caption{Figure (a) illustrates the pre-training process of $Net$ and $Net^{G}$. GAN-based synthetic images are used as an extra training label $c_{K+1}$ apart from known labels $[c_{i,...,K}]$. Figure (b) explains the difference between score calibration in OpenMax and G-OpenMax.}
\label{fig:diff}
\end{figure*}


\subsection{OpenMax}
\label{sec:openmax}
We begin by reviewing the OpenMax open set classifier~\cite{bendale2016towards}. 
It explores the effectiveness of penultimate layer from the deep network to limit open space risk. Like multi-class open set classifier $\varA{W-SVM}$~\cite{scheirer2014probability}, OpenMax holds the property of compact abating probability (CAP) model. 
The learning progress of OpenMax can be summarized into two stages. 
The first stage leverages EVT as a base theory and fit the known classes' post recognition activations to a Weibull distribution. To do so, a base network $Net$ (AlexNet~\cite{krizhevsky2012imagenet} is used in the original paper) is trained as a penultimate activation layer extractor. Mean activation vector (MAV) $MAV = [\mu_{i,...,K}]$ for each class $C = [c_{i,...,K}]$ is computed based on the penultimate activation layer of each sample. Weibull model with hyper-parameters $\varA{p_{c_{i}}=(t_{c_{i}}, \lambda_{c_{i}}, k_{c_{i}})}$ for each class is returned.
The second stage, which is essential in OpenMax, recalibrates each activation of penultimate layer  by estimating Weibull CDF probability on the distance between sample $x_{i}$ and known class' MAV $[\mu_{i,...,K}]$, which servers as the core of the rejection estimation. 
At last, pseudo-probability of unknown class is estimated from known class' activation scores to support explicit rejection.   



%


\subsection{Generative OpenMax (G-OpenMax)}
\label{sec:g-openmax}
Here we describe the overall inference procedure of the proposed method G-OpenMax in Alg.~\ref{alg:g-openmax}. The main difference between vanilla OpenMax and G-OpenMax is shown in Fig.~\ref{fig:diff}. 
G-OpenMax adopts the same two-stage procedure as OpenMax. However, in parallel to training the classifier and Weibull model at the first stage by using known classes $c_{i,...,K}$ only, we train the network $Net^{G}$ with $K+1$ classes (see Fig.~\ref{fig:diff} (a)), where the extra images come from our generator $G$ to represent the unknown class $c_{K+1}$.  
Consequently, the activation vector $v^{G}$ in G-OpenMax has one extra dimension output to represent the $K+1$ class which is regarded as unknown. 
It is noticed that the penultimate layer provides dependent information about how classes are related with each other. Therefore, the extra dimension is able to supply information about MAV distance between known and unknown classes in feature space ($\mu_{i,...,K}$ with respect to $\mu_{K+1}$). 
At the second stage, G-OpenMax follows the same procedure as OpenMax using Weibull CDF probability for score calibration. However, G-OpenMax provides explicit probability estimation of the unknown class (line 6 in Alg.~\ref{alg:g-openmax}). 


%
%
%
%
%

\begin{algorithm}[!t]
\label{alg:g-openmax}
\SetKwInOut{Input}{input}\SetKwInOut{Output}{output}
 \Input{Number of top-score predictions to calibrate, $\alpha$ \\
Pre-set threshold $\epsilon$ }
  \BlankLine
 \Output{ known class probability $Prob_{1,...,K}$ + unknown class probability $Prob_{K+1}$}
 \textbf{Train} $Net$ from known classes $c_{i,...,K}$\;
 \textbf{Train} the generative model $G$\;
 \textbf{Unknown sample selection} using $Net$\;
 \textbf{Train} $Net^{G}$ with known class and unknown class samples\; 
 \textbf{Computing} activation vector $v^{G}(x) = v_{1}(x),...,v_{K+1}(x)$ from $Net^{G}$\;
 \textbf{Fit} Weibull models $p^{G} = [p^{G}_{1,...,K+1}]$\; 
  $argsort(v^{G}(x))$ \;
 \For{c=1,...,$\alpha$}{
  $w_{s(c)}(x) = 1 - \frac{\alpha-c}{\alpha}e^{-(\frac{||v^{G}(x)-t_{s(c)} ||}{\lambda_{s(c)}})^{k_{s(c)}}}$\;
  }
  Recalibrate activation vector:  $v(x)' = v^{G}(x) \odot w(x) $\;
  $Prob = Softmax(v(x)')$\;
  \eIf{$\underset{c}{\max}Prob(y=c|x)<\epsilon$}{
  $pred = unknown class$\;}
  {
  $pred = argmax_{c}P(y=c|x)$\;
  }
 \caption{Generative OpenMax algorithm}
 \label{alg:g-openmax}
\end{algorithm}

\subsection{Understanding Generation of Synthetic Samples} 
\label{sec:syn-samples}
While in theory the open space should be infinitely large, an ideal open set classifier is able to recognize unknown classes come from any distributions. In practice, many recent open set works~\cite{scheirer2013toward,   scheirer2014probability,jain2014multi,bendale2015towards,bendale2016towards} assume that the classes from unknown open space will share some common properties with known classes. To elaborate, if the system is trained on a set of English or Chinese characters, an open set class will probably be another character from the same group. An RGB based colour image is not expected to evaluate the characters trained system. This is a fair assumption to make a first step solution toward the challenging open set problem.  
Thus, in this paper we assume that \textit{open space classes belong to a subspace of the original space}, which includes known classes. In order to preserve it, we also do not consider an unlikely scenario when the test set contains objects from other datasets.
The advantage of building algorithm based on this assumption is that we would turn an open space classification problem into a standard closed set problem by sampling objects from known subspace.  
A natural way to do this is linear interpolation (or linear combination) between objects, belonging to the subspace. However, in the original pixel space this subspace is highly nonlinear, so linear interpolation would not produce any plausible results. For instance, interpolating between photos of two birds will not result in a realistic photo of another bird class.  
However in the latent space, class information is separated from other object characteristics, so the class can be modified independently without affecting them~\cite{perarnau2016invertible}. 
Since all dataset classes are encoded by one of the basis one-hot vectors, we can generate new classes based on the linear combination of those known classes in latent space. 
Moreover, the generated mixed-class samples share a number of common properties with known class images, all of those produced images will have high probability in a given data distribution. Which makes those generated samples good candidates to represent a reasonable open space under our assumption.
Formally, we can assume that existing classes are encoded by one-hot basis vectors $\mathbf{b}\in\mathbb{R}^{N}$, which define the class subspace in the latent space. In this case, we can generate a class mixture vector  $\mathbf{m}\in\mathbb{R}^{N}$ using any distribution $P_{norm}$ (e.g. Normal distribution) by sampling $m_1, \ldots, m_{N-1}$ from $P_{norm}$, and assigning $m_N = 1 - \sum_{i=1}^{N-1} m_i$, such that $\sum_{i=1}^N m_i = 1$ as for the basis vectors $\mathbf{b_{i}}\in\mathbb{R}^{N}$.
The dimensionality of class subspace space (number of classes) is typically much smaller than the dimensionality of the original space (number of pixels), so the generated open set images will be located is a small nonlinear subspace of it as desired.


We employ a modified conditional GAN~\cite{odena2016conditional} to train the generator and then synthesize unknown classes out of it. Note that the algorithm is not GAN-specific, it also generalizes to any generative models, e.g. VAE~\cite{kingma2013auto}.
A typical GAN is composed of a generator $G$ and a discriminator $D$. While $G$ learns to transform random noise to  samples similar to those from the dataset, $D$ learns to differentiate them. In a case of conditional GAN, random noise is fed to $G$ together with a one-hot vector $c \in c_{i,...K}$, which represents a desired class. In this case, discriminator $D$ also learns faster if the input image is supplied together with the class it belongs to. 
Thus, the optimization of a conditional GAN with class labels can be formulated as:
\begin{equation}
\underset{\phi}{\min} \ \underset{\theta}{\max} = E_{x,c \sim p_{data}}[\log D_\theta (x, c)]
+ E_{z \sim P_{z}, c \sim P_{c}}[\log (1-D_\theta (G_\phi (z, c), c))]
\end{equation}
where the generator inputs $z$ and $c$ are the latent variables drawn from their prior distribution $P(z)$ and $P(c)$. Here $\phi$ and $\theta$ denote trainable parameters for $G_\phi$ and $D_\theta$, respectively.

\noindent\textbf{Selection of Generated Samples:} One important step is that we need to select open space samples which do not belong to the subspace of known classes.
For each synthetic sample generated from a mixture of class distribution, the class with the highest value is treated as the ground-truth label. 
We feed all the generated samples $x^{G}_{1,...,N}$ into the pre-trained classifier $Net$, all incorrectly predicted samples are selected as candidates for $c_{k+1}$to train the $NET^{G}$.

\subsection{Implementation Details}
We set hyper-parameters following existing DCGAN work~\cite{radford2015unsupervised}; the GAN is optimized with Adam optimizer ($\beta_{1}=0.5, \beta_{2}=0.999,\epsilon=10^{-8}$) using a learning rate of 2e-4. 
The architectures of both classifier and generator/discriminator of the conditional GAN are provided in Table 1 to 3. 
The generator $G$ and discriminator $D$ takes data $x$ as initial input, and at each linear layer thereafter, the latent representation $c$ ($N=50$) is transformed using a learned linear transformation to the hidden layer dimension and added to the non-linearity input. 
Moreover, to improve the model convergence time, we use soft-labels with 0.9 for training~\cite{salimans2016improved}.

\begin{figure*}[!t]
\begin{center}
\includegraphics[width=0.6\linewidth]{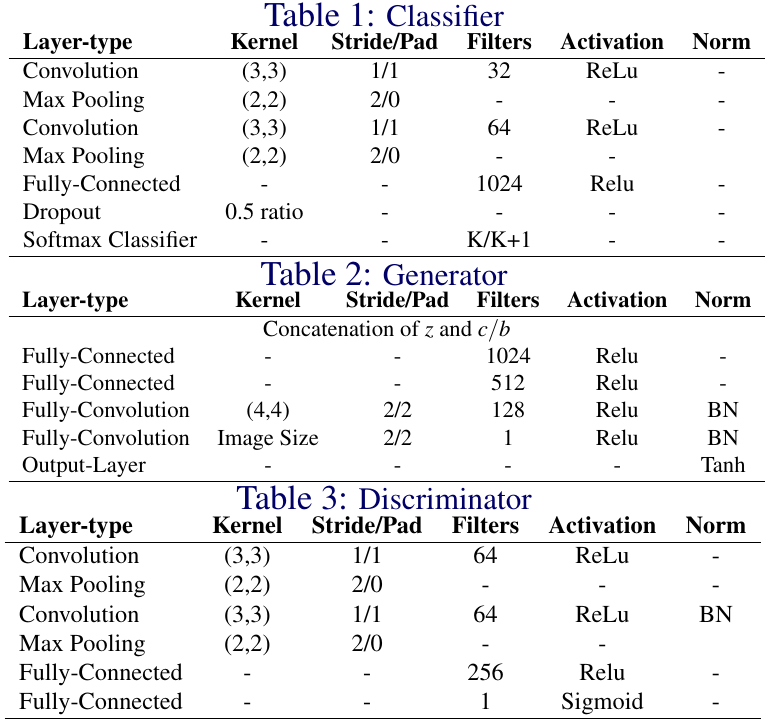}
\end{center}
   \caption*{}
\label{fig:table}
\end{figure*}

%% file: contrib/exp.tex
\section{Experiments}
We perform a thorough comparison of G-OpenMax with OpenMax and its variants on two datasets: MNIST~\cite{lecun1998mnist} and HASYv2~\cite{thoma2017hasyv2}. 
MNIST is a digits dataset of 10 classes from 0 to 9 composed of 70,000 images. 
HASYv2 is the latest dataset of handwritten symbols published in early 2017. It is a more complicated dataset than MNIST in terms of the number of classes (369 classes) and domain of characters (including Arabic and Latin characters). 
However, HASYv2 is not a balanced dataset with respect to the number of samples per class.
In this paper we use only the classes from HASYv2 with no less than 500 samples to ensure enough samples to train a reasonable generator $G$.  
Standard data sets for open set cross-validation is performed for analysis. 
We use 60\% samples of each class for training~\footnote{Given that training a generator $G$ with 60 classes from HASYv2 is difficult to converge, we randomly choose 20 classes from the training pool to train one generator in each iteration. Later on all generated images from different iterations are used together as synthetic samples.}, and 20\% samples are held out for validation. The last 20\% are used for testing. All results are reported as average value from three-fold cross-validation.

\subsection{Main Results}
Following~\cite{scheirer2014probability}, we plot and compare ``openness'' vs F-measure to a few baseline methods in multi-class open set classification. 
The ``openness'' value defines the number of known classes in training and the total number of classes to be recognized in testing. 
\begin{equation}
openness  = 1 - \sqrt{\frac{2N_{Train}}{N_{R}+N_{Test}}}
\end{equation}
where $N_{Train}$ defines the number of training classes, $N_{Test}$ denotes the number of testing classes and $N_{R}$ presents the number of classes to be recognized. 
The setup on MNIST is a replication of the experiment performed in~\cite{jain2014multi}, where 6 classes are hold out in training and the rest 4 classes are used to varying ``openness''. For HASYv2 dataset, we use 60 classes for training and the rest 35 classes are held out for testing as unknown classes.
The results of our method (G-OpenMax) are compared with: 1) \textbf{SoftMax:} Softmax with probability thresholding as unknown class detector. 2) \textbf{OpenMax:} OpenMax described in Sec.~\ref{sec:openmax}. 3) \textbf{G-SoftMax:} Like Softmax, but G-SoftMax employs synthetic samples as unknown class training samples.   

\begin{figure}[!t]
  \centering
  \begin{minipage}{1\columnwidth}
  \centering
  \includegraphics[width=0.48\columnwidth]{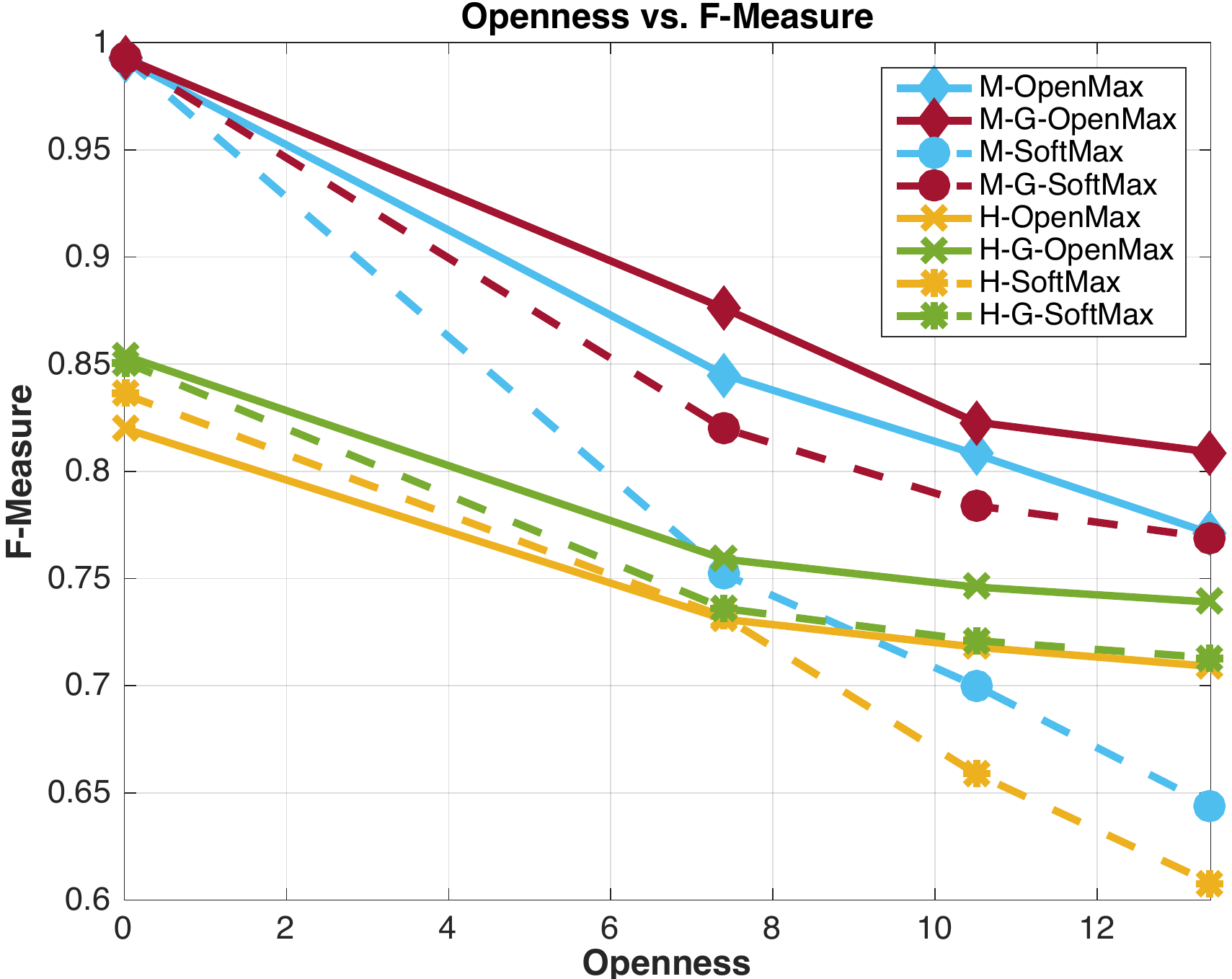}
  \includegraphics[width=0.48\columnwidth]{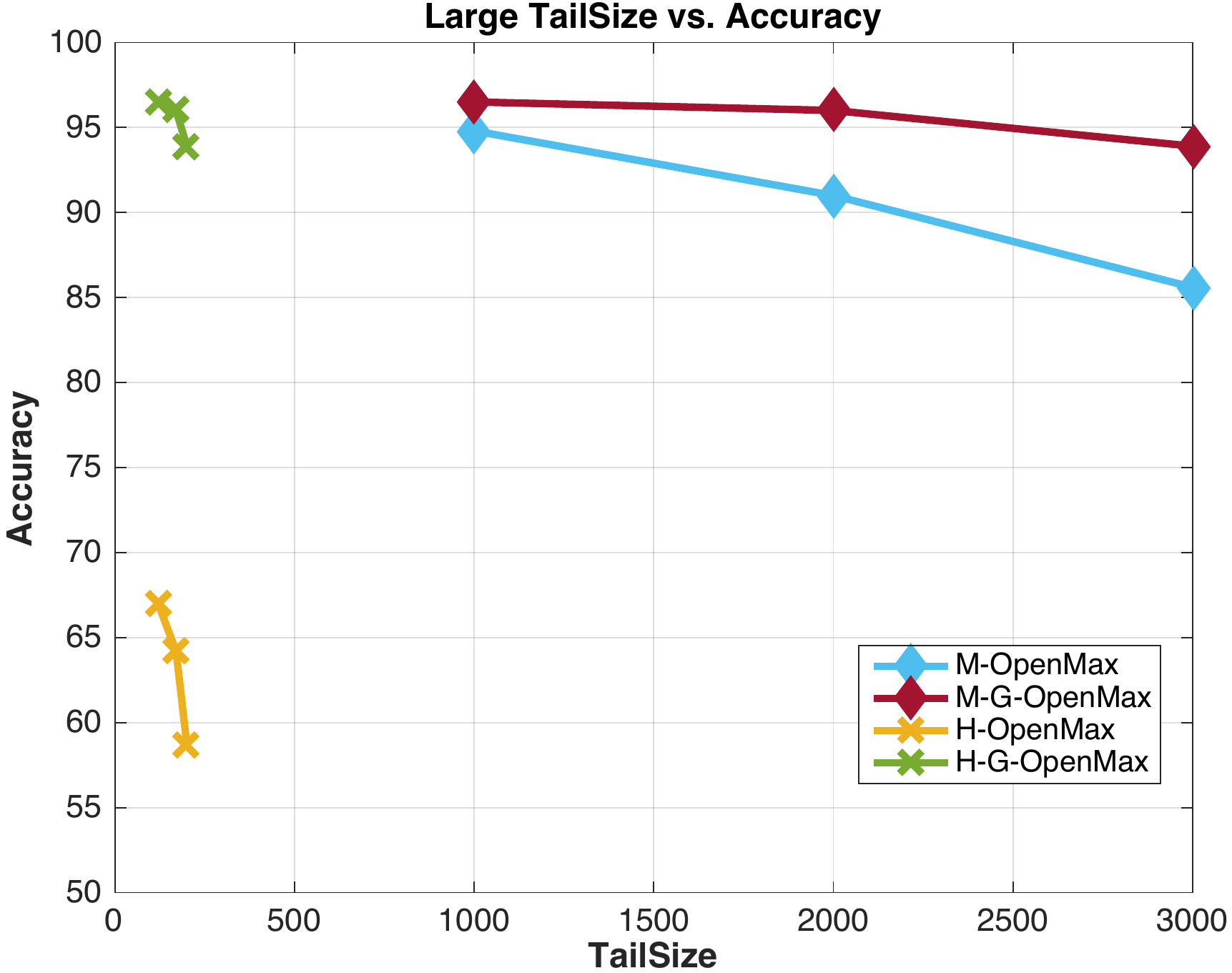}
  \end{minipage}
  \vspace*{5mm}  
  \caption
    {
 	\textbf{Left}: F-measure for multi-class open set recognition on MNIST (\textbf{M}) and HASYv2(\textbf{H}). G-OpenMax maintains high F-measure scores as the openness grows, while vanilla Softmax degrades quickly on MNIST. G-SoftMax is better compared to its baseline SoftMax on both datasets and achieves approximately the same performance as OpenMax on MNIST with openness equal to $13.04$.    
    \textbf{Right}: This graph shows known classes recognition accuracy when using large tail sizes for Weibull model fitting. We observe that G-OpenMax consistently outperforms OpenMax, which leads to the conclusion that G-OpenMax is robust on known samples detection even with large tail size. 
    }
  \label{fig:fmeasure}
\end{figure}

Fig.~\ref{fig:fmeasure} \textit{left} shows open set multi-class recognition F- measure performance for MNIST and HASYv2. 
F-measure for SoftMax degrades rapidly on both test scenarios. On MNIST we see that SoftMax trained with GAN based synthetic samples (G-SoftMax), without score calibration, is competitive against OpenMax. Furthermore, G-Softmax is much better at tolerating increasing openness than normal SoftMax. 
On HASYv2, which has larger number of classes than MNIST, the performance gain of employing GAN samples is expected to be higher than MNIST due to rich variety of generated samples. However, the relatively poor quality of generated mixture samples (see Sec.~\ref{sec:sample}) in HASYv2 hinders the training of G-OpenMax and G-Softmax.   
Overall, the proposed G-OpenMax based approach provides consistent improvement regardless of datasets.

\begin{figure}
  \centering
  \begin{minipage}{1\columnwidth}
  \centering
  \includegraphics[width=0.48\columnwidth]{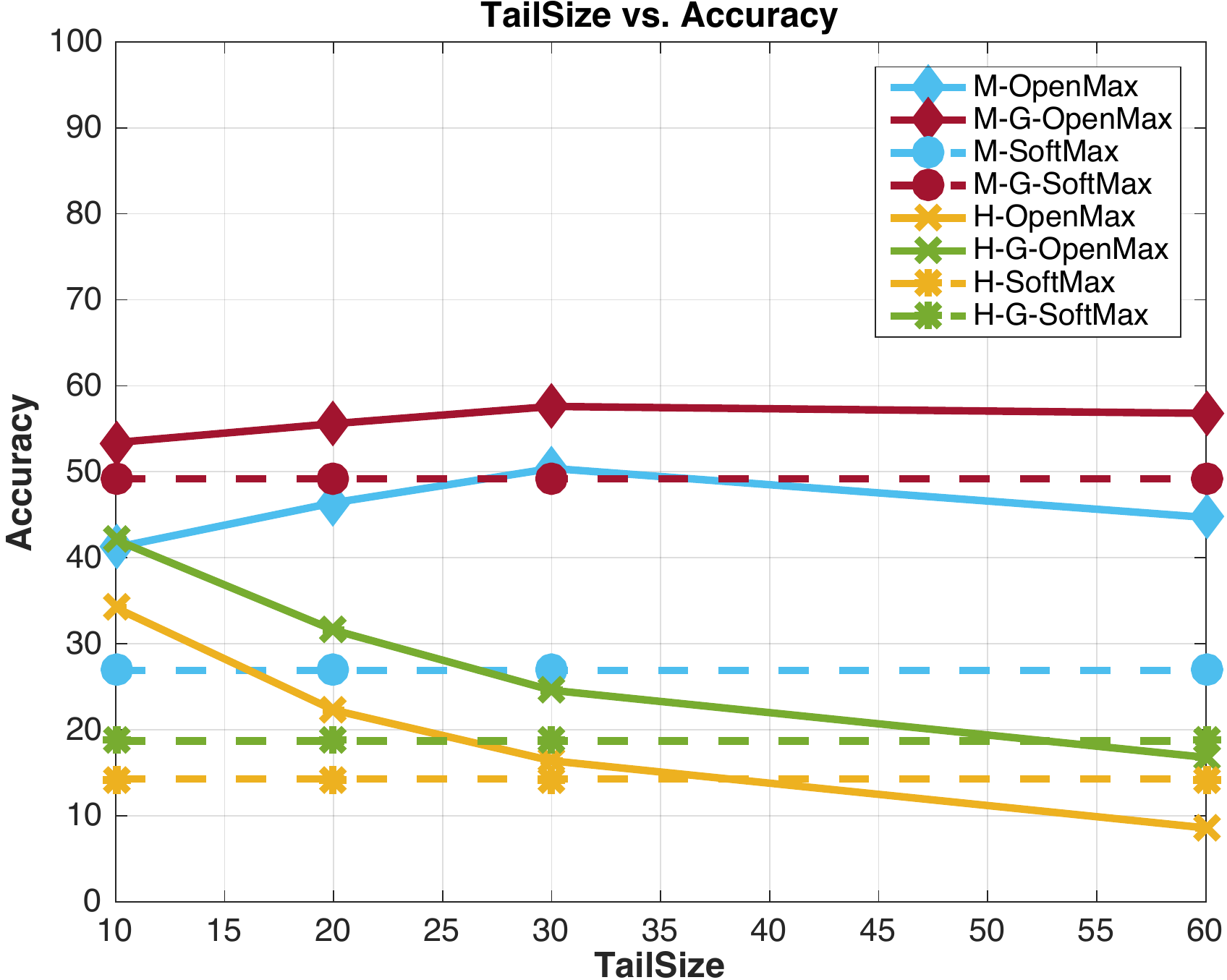}
  \includegraphics[width=0.48\columnwidth]{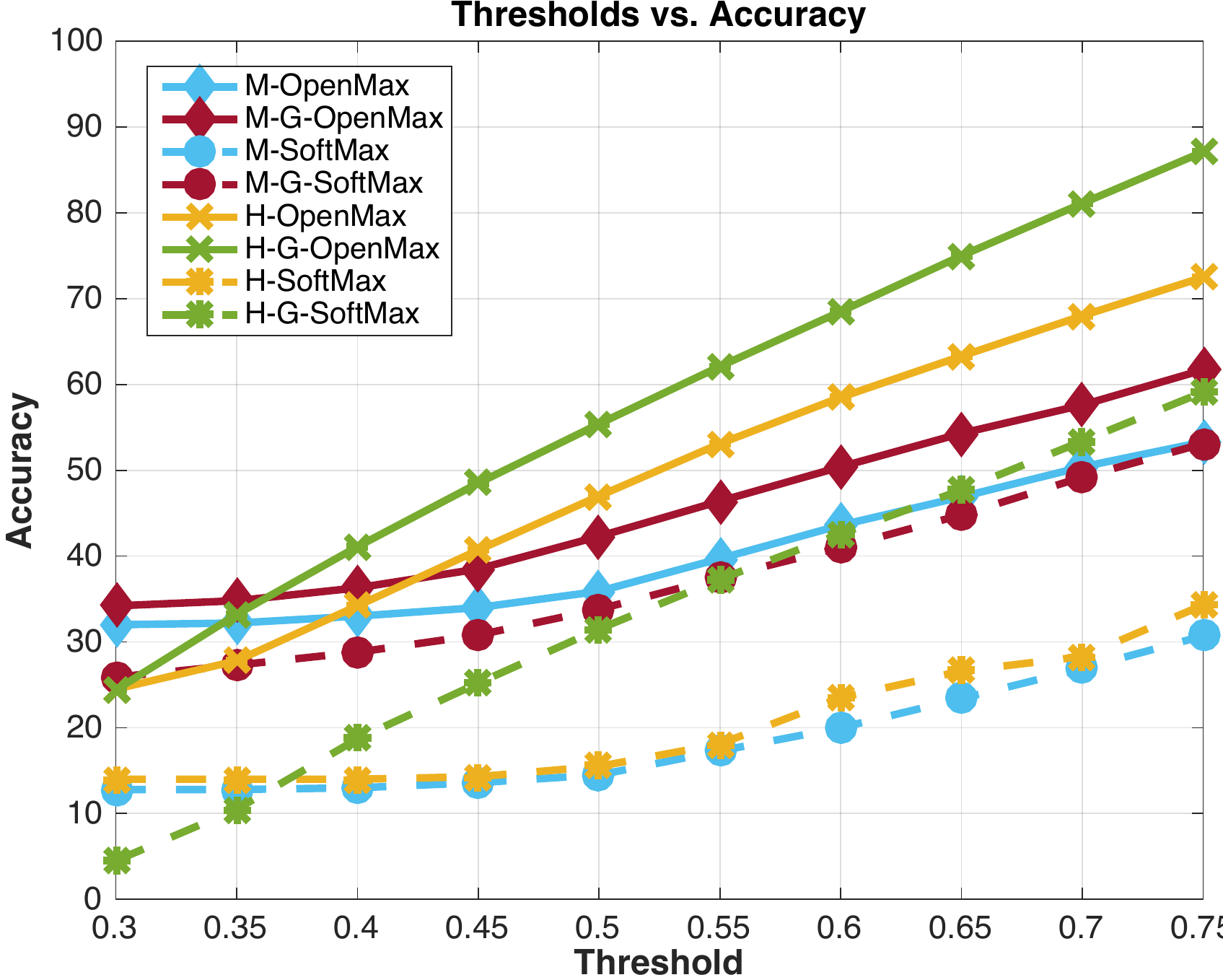}
  \end{minipage}
  \vspace*{5mm}  
  \caption
    {
    \textbf{Left}: This graph shows accuracy on unknown class performance with respect to tail size for open set multi-class recognition. 
    \textbf{Right}: Given different threshold, we report accuracy on unknown class test samples from different methods on MNIST and HASYv2 dataset. 
    }
  \label{fig:tail}
\end{figure}

\subsection{Ablation Experiments}
We run a number of ablation experiments to analyze robustness of different models. The openness evaluated from all experiments in this subsection is $13.04$. To reduce the complexity of running relevant experiments, we use fixed $\alpha$ in those experiments~\footnote{We use $\alpha=2$ for MNIST and $\alpha=4$ for HASYv2.}.  

\noindent\textbf{Tail Sizes for Score Calibration:}
Fig.~\ref{fig:tail} \textit{left} presents the results of varying tail size and evaluating the unknown class accuracy with its optimal threshold. 
We observe that performance on MNIST continues to improve upto tail size of 60. However, large tail size does not work well on HASYv2 dataset. This may attribute to the difficulty of generating meaningful samples from HASYv2 dataset since the complexity and large variations of known classes. 

\noindent\textbf{Large Tail Size vs. Performance:}
In this part, we tried training EVT model (OpenMax and G-OpenMax) with extremely large tail size. 
Although large tail size tends to increase the EVT model rejections for unknown samples, it also increases rejection for known classes~\cite{bendale2016towards}.  
According to our observations from Fig.~\ref{fig:fmeasure} \textit{left}, G-OpenMax is able to maintain the same overall performance even with large tail size. 

\noindent\textbf{Threshold vs. Performance:}
We evaluated performance of different models on unknown test samples with optimal tail size\footnote{tail sizes are selected according to Fig.~\ref{fig:fmeasure} \textit{left}}. 
Our experiments results (see Fig.~\ref{fig:fmeasure} \textit{right}) highlights that the methods trained with GAN samples consistently outperforms methods without employing GAN samples.
G-OpenMax is nearly (+10\%) absolute performance improvement over OpenMax with optimal threshold. 

\subsection{Qualitative Analysis}
\label{sec:sample}
\begin{figure*}
\begin{center}
\includegraphics[width=1\linewidth]{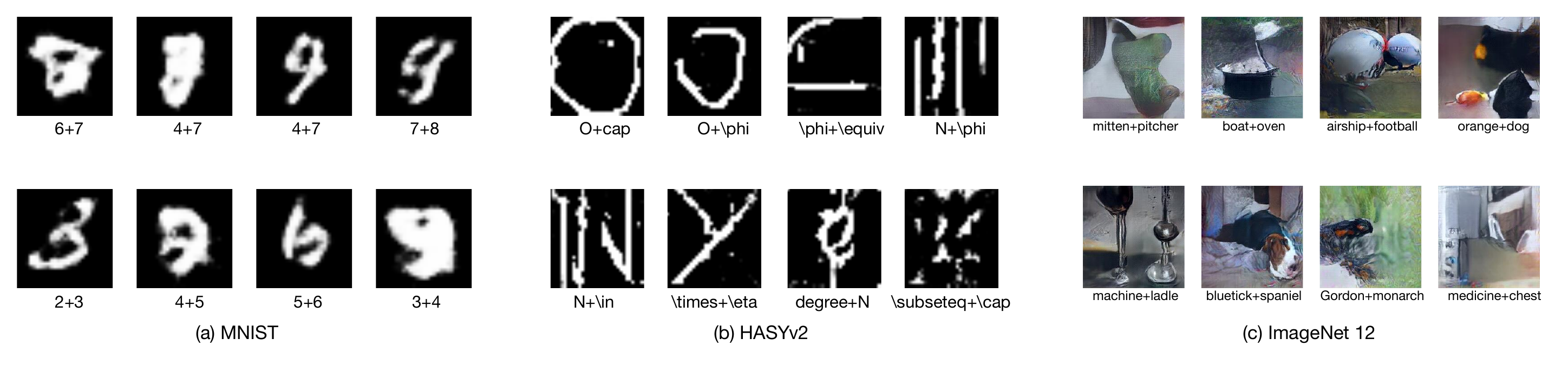}
\end{center}
   \caption{Figure (a) illustrates the synthesis samples of unknown classes. Training classes include: 2-7 from MNIST dataset. It is quite surprising to see that the generator is able to generate outputs like digits 8 and 9 (variants from digits class 3 and 4). Figure (b) shows mixture samples from HASYv2 dataset. Random dots are occasionally being seen from the background. Figure (c) illustrates mixture samples learned from natural image samples. Highest two class activations are displayed at the bottom of the image.}
\label{fig:mix}
\end{figure*}

It is helpful to visualize the mixture samples generated by GAN to get an insight about the classes used as training samples for G-SoftMax and G-OpenMax. Fig.\ref{fig:mix} (a) and (b) demonstrate mixture examples generated from MNIST, HASYv2 and ImageNet12 respectively. 
Mixture samples from MNIST show some distinct attributes of digits, such as stroke continuity and orientation etc. 
Mixture samples from HASYv2 are sometimes hard for human to understand (second row last sample). The main reason may attributes to the complexity of the HASYv2 dataset since it consists of English alphabets, Greek letters etc.

A natural extension of our method is to perform experiments on the natural image dataset. 
We trained the generator $G$ on ImageNet12 and made several observations: 
1) Several classes blend into unrealistic type of object in terms of colour and global structure. 
2) Mixture model tend to collapse into a single class appearance output. 
However, we perfomed the same experiment as in~\cite{bendale2016towards} and figured out there are no obvious performance improvement by using G-OpenMax over the natural image setting. 
A more general reason might be that the generated images are not plausible with respect to the training classes in order to be good candidates to represent unknown classes from open space. 
More specifically, unlike the MNIST and HASYv2 dataset, ImageNet classes enjoy large variety, therefore not many common features are encoded in the learned latent space. Since our method is based on disentangling class information from other object features, the results it produces in this case do not differ much from interpolation in the pixel space.


%


%% file: contrib/conclusion.tex
\section{Conclusion}
In this paper, we presented a novel method for multi-class open set classification. 
The proposed method G-OpenMax extends OpenMax by applying generative adversarial network and is able to provide explicit probability estimation and visualization of unknown classes.
Based on the reasonable assumption for open space modelling, which assumes that all classes share common features. Results demonstrate that when this assumption holds, results are indeed better. Moreover, we have visualised the objects and demonstrated that they look plausible with respect to training classes. 
However, there still remains some unanswered questions to be addressed such as how the complexity of mixture classes affects the performance of G-OpenMax? 
How to solve the open set classification for natural images?
In the future work, we will explore those directions.
\footnote{$^{\star}$Equal contribution of this paper.}

%% file: bmvc_final.bbl
\begin{thebibliography}{29}
\providecommand{\natexlab}[1]{#1}
\providecommand{\url}[1]{\texttt{#1}}
\expandafter\ifx\csname urlstyle\endcsname\relax
  \providecommand{\doi}[1]{doi: #1}\else
  \providecommand{\doi}{doi: \begingroup \urlstyle{rm}\Url}\fi

\bibitem[Arjovsky et~al.(2017)Arjovsky, Chintala, and
  Bottou]{arjovsky2017wasserstein}
Martin Arjovsky, Soumith Chintala, and L{\'e}on Bottou.
\newblock Wasserstein gan.
\newblock \emph{arXiv preprint arXiv:1701.07875}, 2017.

\bibitem[Bendale and Boult(2015)]{bendale2015towards}
Abhijit Bendale and Terrance Boult.
\newblock Towards open world recognition.
\newblock In \emph{CVPR}, 2015.

\bibitem[Bendale and Boult(2016)]{bendale2016towards}
Abhijit Bendale and Terrance~E Boult.
\newblock Towards open set deep networks.
\newblock In \emph{CVPR}, 2016.

\bibitem[Creswell and Bharath(2016)]{creswell2016adversarial}
Antonia Creswell and Anil~Anthony Bharath.
\newblock Adversarial training for sketch retrieval.
\newblock In \emph{ECCV}. Springer, 2016.

\bibitem[Deng et~al.(2009)Deng, Dong, Socher, Li, Li, and Fei-Fei]{Deng2009}
Jia Deng, Wei Dong, Richard Socher, Li-Jia Li, Kai Li, and Li~Fei-Fei.
\newblock Imagenet: A large-scale hierarchical image database.
\newblock In \emph{CVPR}, 2009.

\bibitem[Finn et~al.(2016)Finn, Goodfellow, and Levine]{finn2016unsupervised}
Chelsea Finn, Ian Goodfellow, and Sergey Levine.
\newblock Unsupervised learning for physical interaction through video
  prediction.
\newblock In \emph{NIPS}, pages 64--72, 2016.

\bibitem[Goodfellow et~al.(2014)Goodfellow, Pouget-Abadie, Mirza, Xu,
  Warde-Farley, Ozair, Courville, and Bengio]{goodfellow2014generative}
Ian Goodfellow, Jean Pouget-Abadie, Mehdi Mirza, Bing Xu, David Warde-Farley,
  Sherjil Ozair, Aaron Courville, and Yoshua Bengio.
\newblock Generative adversarial nets.
\newblock In \emph{NIPS}, 2014.

\bibitem[He et~al.(2016)He, Zhang, Ren, and Sun]{he2015deep}
Kaiming He, Xiangyu Zhang, Shaoqing Ren, and Jian Sun.
\newblock Deep residual learning for image recognition.
\newblock \emph{in CVPR}, 2016.

\bibitem[He et~al.(2017)He, Gkioxari, Doll{\'a}r, and Girshick]{he2017mask}
Kaiming He, Georgia Gkioxari, Piotr Doll{\'a}r, and Ross Girshick.
\newblock Mask r-cnn.
\newblock \emph{arXiv preprint arXiv:1703.06870}, 2017.

\bibitem[Isola et~al.(2016)Isola, Zhu, Zhou, and Efros]{isola2016image}
Phillip Isola, Jun-Yan Zhu, Tinghui Zhou, and Alexei~A Efros.
\newblock Image-to-image translation with conditional adversarial networks.
\newblock \emph{arXiv preprint arXiv:1611.07004}, 2016.

\bibitem[Jain et~al.(2014)Jain, Scheirer, and Boult]{jain2014multi}
Lalit~P Jain, Walter~J Scheirer, and Terrance~E Boult.
\newblock Multi-class open set recognition using probability of inclusion.
\newblock In \emph{ECCV}, 2014.

\bibitem[Kingma and Welling(2013)]{kingma2013auto}
Diederik~P Kingma and Max Welling.
\newblock Auto-encoding variational bayes.
\newblock \emph{arXiv preprint arXiv:1312.6114}, 2013.

\bibitem[Krizhevsky et~al.(2012)Krizhevsky, Sutskever, and
  Hinton]{krizhevsky2012imagenet}
Alex Krizhevsky, Ilya Sutskever, and Geoff Hinton.
\newblock Imagenet classification with deep convolutional neural networks.
\newblock In \emph{NIPS}, 2012.

\bibitem[LeCun et~al.(1998)LeCun, Cortes, and Burges]{lecun1998mnist}
Yann LeCun, Corinna Cortes, and Christopher~JC Burges.
\newblock The mnist database of handwritten digits, 1998.

\bibitem[Ledig et~al.(2016)Ledig, Theis, Husz{\'a}r, Caballero, Cunningham,
  Acosta, Aitken, Tejani, Totz, Wang, et~al.]{ledig2016photo}
Christian Ledig, Lucas Theis, Ferenc Husz{\'a}r, Jose Caballero, Andrew
  Cunningham, Alejandro Acosta, Andrew Aitken, Alykhan Tejani, Johannes Totz,
  Zehan Wang, et~al.
\newblock Photo-realistic single image super-resolution using a generative
  adversarial network.
\newblock \emph{arXiv preprint arXiv:1609.04802}, 2016.

\bibitem[May(1988)]{may1988many}
Robert~M May.
\newblock How many species are there on earth?
\newblock \emph{Science}, 241\penalty0 (4872):\penalty0 1441, 1988.

\bibitem[Niculescu-Mizil and Caruana(2005)]{niculescu2005predicting}
Alexandru Niculescu-Mizil and Rich Caruana.
\newblock Predicting good probabilities with supervised learning.
\newblock In \emph{ICML}, pages 625--632. ACM, 2005.

\bibitem[Odena et~al.(2016)Odena, Olah, and Shlens]{odena2016conditional}
Augustus Odena, Christopher Olah, and Jonathon Shlens.
\newblock Conditional image synthesis with auxiliary classifier gans.
\newblock \emph{arXiv preprint arXiv:1610.09585}, 2016.

\bibitem[Perarnau et~al.(2016)Perarnau, van~de Weijer, Raducanu, and
  {\'A}lvarez]{perarnau2016invertible}
Guim Perarnau, Joost van~de Weijer, Bogdan Raducanu, and Jose~M {\'A}lvarez.
\newblock Invertible conditional gans for image editing.
\newblock \emph{arXiv preprint arXiv:1611.06355}, 2016.

\bibitem[Radford et~al.(2016)Radford, Metz, and
  Chintala]{radford2015unsupervised}
Alec Radford, Luke Metz, and Soumith Chintala.
\newblock Unsupervised representation learning with deep convolutional
  generative adversarial networks.
\newblock \emph{ICLR}, 2016.

\bibitem[Russakovsky et~al.(2015)Russakovsky, Deng, Su, Krause, Satheesh, Ma,
  Huang, Karpathy, Khosla, Bernstein, et~al.]{russakovsky2015imagenet}
Olga Russakovsky, Jia Deng, Hao Su, Jonathan Krause, Sanjeev Satheesh, Sean Ma,
  Zhiheng Huang, Andrej Karpathy, Aditya Khosla, Michael Bernstein, et~al.
\newblock Imagenet large scale visual recognition challenge.
\newblock \emph{IJCV}, 115\penalty0 (3):\penalty0 211--252, 2015.

\bibitem[Salimans et~al.(2016)Salimans, Goodfellow, Zaremba, Cheung, Radford,
  and Chen]{salimans2016improved}
Tim Salimans, Ian Goodfellow, Wojciech Zaremba, Vicki Cheung, Alec Radford, and
  Xi~Chen.
\newblock Improved techniques for training gans.
\newblock In \emph{NIPS}, 2016.

\bibitem[Scheirer et~al.(2011)Scheirer, Rocha, Micheals, and
  Boult]{scheirer2011meta}
Walter~J Scheirer, Anderson Rocha, Ross~J Micheals, and Terrance~E Boult.
\newblock Meta-recognition: The theory and practice of recognition score
  analysis.
\newblock \emph{PAMI}, 33\penalty0 (8):\penalty0 1689--1695, 2011.

\bibitem[Scheirer et~al.(2013)Scheirer, de~Rezende~Rocha, Sapkota, and
  Boult]{scheirer2013toward}
Walter~J Scheirer, Anderson de~Rezende~Rocha, Archana Sapkota, and Terrance~E
  Boult.
\newblock Toward open set recognition.
\newblock \emph{PAMI}, 35\penalty0 (7):\penalty0 1757--1772, 2013.

\bibitem[Scheirer et~al.(2014)Scheirer, Jain, and
  Boult]{scheirer2014probability}
Walter~J Scheirer, Lalit~P Jain, and Terrance~E Boult.
\newblock Probability models for open set recognition.
\newblock \emph{PAMI}, 36\penalty0 (11):\penalty0 2317--2324, 2014.

\bibitem[Smith(1990)]{smith1990extreme}
Richard~L Smith.
\newblock Extreme value theory.
\newblock \emph{Handbook of applicable mathematics}, 7:\penalty0 437--471,
  1990.

\bibitem[Smola(2000)]{smola2000advances}
Alexander~J Smola.
\newblock \emph{Advances in large margin classifiers}.
\newblock MIT press, 2000.

\bibitem[Szegedy et~al.(2016)Szegedy, Ioffe, Vanhoucke, and
  Alemi]{szegedy2016inception}
Christian Szegedy, Sergey Ioffe, Vincent Vanhoucke, and Alex Alemi.
\newblock Inception-v4, inception-resnet and the impact of residual connections
  on learning.
\newblock \emph{arXiv preprint arXiv:1602.07261}, 2016.

\bibitem[Thoma(2017)]{thoma2017hasyv2}
Martin Thoma.
\newblock The hasyv2 dataset.
\newblock \emph{arXiv preprint arXiv:1701.08380}, 2017.

\end{thebibliography}
